\documentclass[twoside,leqno,twocolumn]{article}

\usepackage[letterpaper]{geometry}

\usepackage{ltexpprt}
\usepackage{amsmath,amsfonts,amssymb,bm}
\usepackage{hyperref}
\usepackage{url}
\usepackage{booktabs}
\usepackage{graphicx}
\usepackage[english]{babel}
\usepackage{stmaryrd}
\usepackage{multicol}
\usepackage{multirow}
\newtheorem{definition}{DEFINITION}
\newcommand{\nop}[1]{}
\newcommand{\ACTS}{\textbf{ACTS}}

\begin{document}

\title{\Large Inter-Series Attention Model for COVID-19 Forecasting}
\author{Xiaoyong Jin\thanks{University of California, Santa Barbara, CA, USA. Email:\{x\_jin,xyan,yuxiangw\}@cs.ucsb.edu} \and Yu-Xiang Wang\footnotemark[1] \and Xifeng Yan\footnotemark[1]}

\date{}

\maketitle

% Copyright Statement
% When submitting your final paper to a SIAM proceedings, it is requested that you include 
% the appropriate copyright in the footer of the paper.  The copyright added should be 
% consistent with the copyright selected on the copyright form submitted with the paper.
% Please note that "20XX" should be changed to the year of the meeting.

% Default Copyright Statement
\fancyfoot[R]{\scriptsize{Copyright \textcopyright\ 2020 by SIAM\\
Unauthorized reproduction of this article is prohibited}}

% Depending on which copyright you agree to when you sign the copyright form, the copyright 
% can be changed to one of the following after commenting out the default copyright statement
% above.

%\fancyfoot[R]{\scriptsize{Copyright \textcopyright\ 20XX\\
%Copyright for this paper is retained by authors}}

%\fancyfoot[R]{\scriptsize{Copyright \textcopyright\ 20XX\\
%Copyright retained by principal author's organization}}

%\pagenumbering{arabic}
%\setcounter{page}{1}%Leave this line commented out.

\begin{abstract} \small\baselineskip=9pt 
COVID-19 pandemic has an unprecedented impact all over the world since early 2020. During this public health crisis, reliable  forecasting of the disease becomes critical for resource allocation and administrative planning.  The results from compartmental models such as SIR and SEIR are popularly referred by CDC and news media. With more and more COVID-19 data becoming available, we examine the following question: Can a direct  data-driven approach without modeling the disease spreading dynamics outperform the well referred compartmental models and their variants? In this paper, we show the possibility. It is observed that as COVID-19 spreads at different speed and scale in different geographic regions, it is highly likely that similar progression patterns are shared among these regions within different time periods. This intuition lead us to develop a new neural forecasting model, called Attention Crossing Time Series (\textbf{ACTS}), that makes forecasts via comparing patterns across time series obtained from multiple regions. The attention mechanism originally developed for natural language processing can be leveraged and generalized to materialize this idea.  Among 13 out
of 18 testings including forecasting newly confirmed cases, hospitalizations and deaths, \textbf{ACTS} outperforms all the leading COVID-19 forecasters highlighted by CDC. 

\nop{Furthermore, we show that the learned pattern embeddings and attention maps provide insight for epidemiological analysis.}
\end{abstract}
\textbf{keywords}: COVID-19, Time Series Forecasting, Attention, Detrending

\section{Introduction}

The Coronavirus disease 2019 (COVID‑19) has been impacting the human society since early 2020. At the time of this writing, it is an ongoing public health crisis in over 187 countries and territories around the world, with more than 30 million confirmed cases, and a growing death toll exceeding $1,000,000$. During this crisis, reliable forecasting of COVID-19 cases becomes important as it will help
(1) healthcare institutes to allocate sufficient supply and resources,
(2) policy-makers to consider new and further administrative interventions,
(3) general public to be aware of the situation and to follow rules against the epidemic. 
Therefore, the Center for Disease Control and Prevention (CDC) has been actively collecting and publishing data about confirmed cases, hospitalization and deaths related to COVID-19, and hosting forecasting results in the coming weeks. 

The US has been suffering the most severe loss from the pandemic, in which more than $200,000$ lives were lost. To encourage and to bring together efforts of COVID-19 modeling, CDC has launched a forecasting challenge\footnote{\url{https://www.cdc.gov/coronavirus/2019-ncov/cases-updates/forecasting.html}}. It calls for models that give predictions of the next $4$ weeks on a daily or weekly basis.  Besides COVID-19 data, other kinds of data such as demographic data, mobility data and intervention policies are also encouraged to be used in predictions.

Epidemic forecasting is regarded as a challenging task for a long time, for which many methods have been developed.  They can be roughly categorized into two classes:
\begin{enumerate}
    \item \textbf{Compartmental models} These models explicitly compartmentalize the population in groups based on their status of infection and recovery, and simulate the transmission process using differential equations. As of today, most of the CDC-featured forecasting methods fall into this category. Examples includes \cite{arik2020interpretable, pei2020initial, yang2020covid} that are built upon classic SIR or SEIR models \cite{harko2014exact}.  Compartmental models describe disease spreading dynamics; however, it is quite hard to determine parameters in these models as they are influenced by many uncontrollable and dynamically changing factors.
    \item \textbf{Statistical models} This type of methods fits the data to regression models directly, such as \cite{altieri2020curating, murray2020forecasting, woody2020projections}. While they are more flexible in processing real data compared to compartmental models, they often assume a simplified model class such as generalized linear models \cite{altieri2020curating}, or require sophisticated hand-crafted features from additional, and possibly proprietary, data sources \cite{woody2020projections}. 
\end{enumerate}
The forecasting of COVID-19 is even harder as various constantly changing factors, such as virus characteristics, social and cultural distinctions, public attitudes and behaviors, intervention policies and healthcare preparation, influence the contagious rate and death rate significantly. Will there be a better alternative that is solely data-driven without any assumptions about the underlying disease propagation mechanisms? In particular, we experimented a set leading neural forecasters \cite{li2019enhancing, lim2019temporal, rodriguez2020deepcovid}, but none of them gave the best result.

\begin{figure}
    \centering
    \includegraphics[scale=0.45]{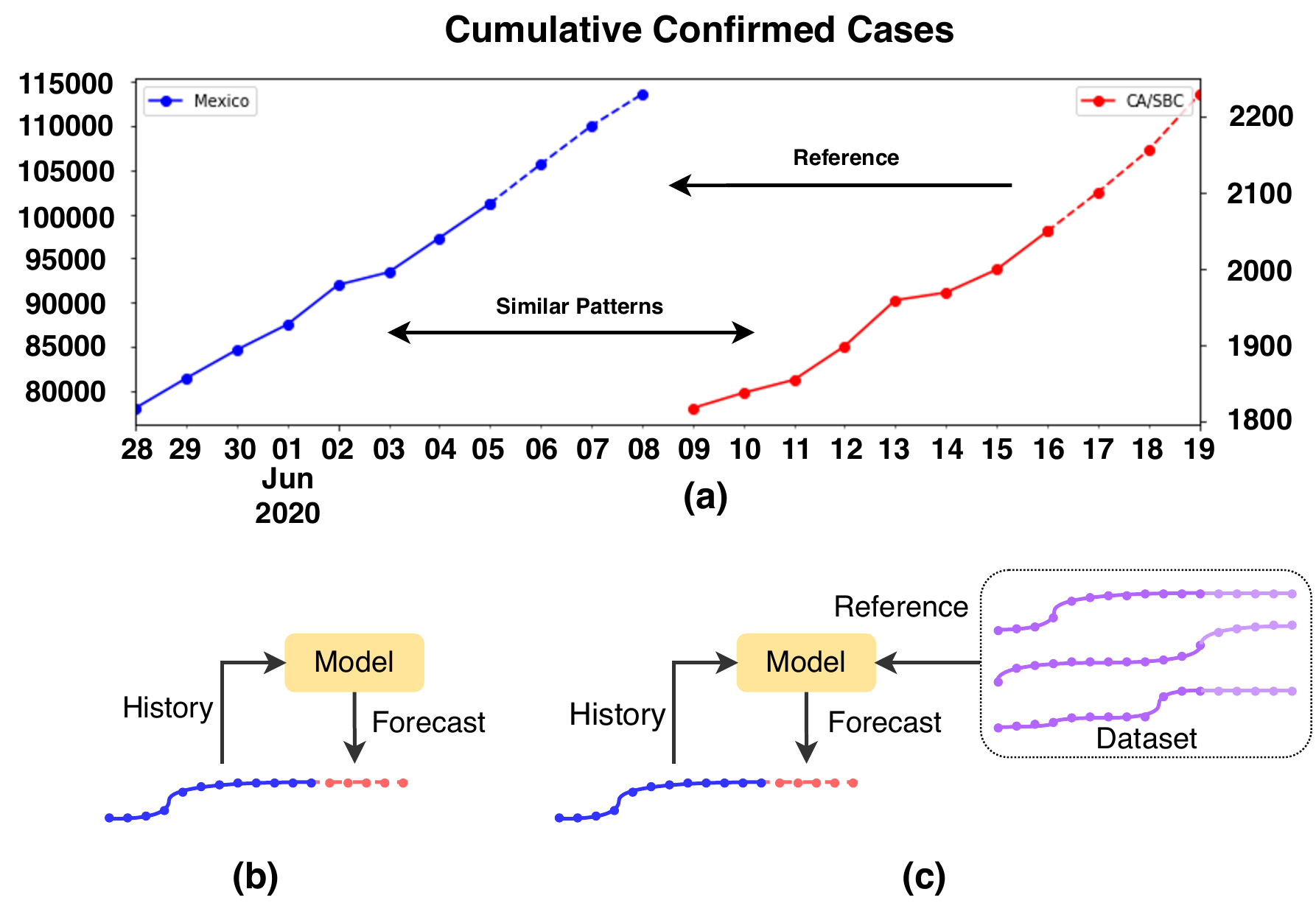}
    \caption{\textbf{(a)} A similar growth pattern of confirmed cases in Santa Barbara County, California, in mid June is  observed in Mexico in late May and early June. \textbf{(b)} Conventional auto-regressive forecasting model. \textbf{(c)} The proposed inter-series forecasting model (\ACTS).}
    \label{fig:concept}
\end{figure}

Since the deep models are originally designed for sufficiently long time series with hundreds of points, the scarce historical data in this task might be the reason of their failures. A natural alternative is to exploit other time series in the dataset if they reveals similar dynamics. Fortunately, even if any two regions present different disease curves over long term, it is likely to find short periods in which different regions sharing similar patterns. Figure \ref{fig:concept}(a) shows surprisingly that the growth pattern of confirmed cases in Santa Barbara county, California, is highly similar to that in Mexico $11$ days ago even though at different scales. Moreover, the further growth in Santa Barbara is also close to that within the corresponding time period in Mexico. In light of this key observation, it is intuitively possible to do better forecasting for Santa Barbara by referring to Mexico in this specific time window via proper transformations.
%A careful examination reveals that as the number of COVID-19 cases changes over time, it is hard to find two regions share the same progression pattern over a long period.  However, it is likely to find a period in the past in which two regions share the same pattern.  

%Given a time series dataset $D=\{\bm{x}^i\}$, in the training stage, one can build a forecasting model $M$ based on $D$. In the inference (forecasting) stage, traditionally, one can use $M$ to predict the future data points in $\bm{x}^i$ without looking back into the historical data of $\bm{x}^j, j \neq i$ any more. In this work, we propose a paradigm change: We will use $M$ and refer to all the data in $D$ to do forecasting in the inference stage. Certainly, the model training has to be adjusted to fit this new paradigm. 
Based on this intuition, we propose to generalize the conventional auto-regressive forecasting to a novel paradigm: besides the local historical data, we also refer to the past reports in all other regions simultaneously in forecasting.
Figures \ref{fig:concept}(b) and (c) illustrate the the fundamental difference between the two paradigms. With time series data of COVID-19 from various locations accumulating over time, we are able to deliver a model outperforming the existing methods by inter-series modeling. Note that unlike other cross-location epidemic forecasters such as \cite{deng2019graph}, only certain time periods rather than the entire time series from other regions will be referred to. 

\nop{We gave new deep learning models a try including  \cite{rodriguez2020deepcovid, li2019enhancing, lim2019temporal}. But they did not give a promising result according to our experiments. , because there is limited data about COVID-19 as it is a newly emerging disease, and the spreading patterns are too diverse and complex for a single model to capture.}

In order to make the proposed paradigm work, it is critical to find small segments in reference time series that exhibit similarity with target time series. It turns out that the attention mechanism originated in natural language process \cite{vaswani2017attention} is a good choice for pattern matching. Moreover, it is found that solely applying attention does not work the best as the embedded small segments do not contain long-term trends that are not directly comparable. We filter out these trends and introduce a normalization step so that the small segments can be matched at a consistent scale. In the end, we put all of these components together and achieve global optimum by joint training. Our new model called \textbf{ACTS} (Attention Crossing multiple Time Series), is able to outperform leading forecasters hosted at CDC.   

Our main contributions are summarized as follows:
\begin{itemize}
    \item We develop a new paradigm that leverages inter-series similarity to improve COVID-19 forecasting. Our method makes no assumption about epidemiological dynamics. 
    \item We extend the attention mechanism to capture inter-series similarity in time series data. Trend filtering is also introduced to complement the attention-based framework and it can be trained jointly to maximize the performance. 
    \item In comparison with a wide range of existing forecasters, the outstanding performance of \ACTS\ is demonstrated on COVID-19 data.
\end{itemize}

%Our main contribution is the development of the new inference paradigm in time series forecasting and demonstrate its outstanding performance on COVID-19 forecasting without assuming disease spreading models and parameters associated with these models.

%In this paper, we develop a purely data-driven model for COVID-19 forecasting based on neural networks, specifically, attention mechanism. By directly learning embeddings of incidence curve in different regions and different time periods, we are able to find similar growth patterns across time series via attention. As a result, we are able to produce better predictions based on learned cross-spatial patterns rather than simple auto-regression or hand-crafted spatial features. To demonstrate the effectiveness of our design. we compare our model with several recent COVID-19 forecasting models and state-of-the-art deep forecasting techniques and observe encouraging results. Furthermore, we show that the learned pattern embeddings and attention maps provide insight to explain questions that epidemiologists may have.

\section{Related Work}
There has been a large body of work focusing on epidemic forecasting. To incorporate domain knowledge, mechanistic models \cite{lessler2016trends, lessler2016mechanistic, zhang2017forecasting} has been favored since they often consider various factors such as epidemiological and social properties, and they make forecasts based on simulation. Moreover, geographic information can also be incorporated into the mechanistic models to better illustrate the spreading process of an infectious disease \cite{balcan2009multiscale, balcan2010modeling}. These models have excellent interpretability but often fail to fit real observed data due to their rigid and over-simplified assumptions without careful calibration.

On the other hand, statistical methods explicitly fit historical data to a statistical model and use it to obtain predictions by extrapolation \cite{brooks2015flexible, chakraborty2014forecasting}. For example, \cite{ray2017infectious} relies on kernel density estimation, \cite{martinez2011sarima} uses seasonal ARIMA, \cite{yang2014comparison} chooses particle filtering and \cite{zimmerinfluenza2020} employs Gaussian process regression. These methods are either too simple or require laborious feature engineering. Hence, various deep learning techniques are also introduced to forecast disease spreading, such as \cite{wu2018deep, wang2019defsi, deng2019graph, chimmula2020time, huang2020novel, tian2020covid, tian2020forecasting, ramchandani2020deepcovidnet, gao2020stan}. They use deep neural networks to extract complex temporal patterns from historical data and a selected set of additional features. \cite{deng2019graph, gao2020stan} are conceptually closer to our model, both of which employ attention mechanism to compare encoded temporal patterns across multiple locations. However, they require a fixed graph structure with geographic information and produce a similarity score between locations that is independent of time. Instead, in our model we generate embeddings of dynamical patterns for attention over both spatial and temporal dimensions so that the generated attention map are temporally dynamical and free from any predefined geographic structures.

\begin{table}[t]
    \centering
    \begin{tabular}{l|l}
    \toprule
        symbol &  interpretation\\
    \midrule
        $\bm{x}^i_t$ & The value at time t in location $i$.\\
        $\bm{x}^i_{s:t}$ & The time series from $s$ to $t$ in location $i$\\
        $\bm{W}_\cdot$ & Parameter matrices to be learned.\\
        $[\bm{a};\bm{b}]$ & The concatenation of $\bm{a}$ and $\bm{b}$.\\
        $\langle\bm{a},\bm{b}\rangle$ & Inner product of $\bm{a}$ and $\bm{b}$.\\
        $\llbracket s, t\rrbracket$ & Consecutive index set $s,s+1,\cdots,t$\\
    \bottomrule
    \end{tabular}
    \caption{Used notations}
    \label{tab:symbols}
\end{table}

\section{Problem Statement}
\nop{In this section, we introduce the COVID-19 forecasting problem and formalize it with customized notations. }
In COVID-19 forecasting, there are three types of incidences, namely confirmed cases, hospitalizations and deaths, to be predicted. The historical data is reported on a daily basis, and we will predict them for the coming weeks. Table \ref{tab:symbols} summarizes the notations we use in the following sections. Note that throughout the paper, terms ``location" and ``region" will be used interchangeably. Problem definition is formulated as follows.  

\begin{definition}\textbf{Incidence Time Series}
We denote by $\bm{x}^i_t$ the reported value of a certain type of incidence data at date $t$ and location $i$, for $t=1,2,\cdots, T$ and $i=1,2,\cdots,N$. Hence, the incidence time series of location $i$ denoted by $\bm{x}^i_{1:T}$. $\bm{x}^i_{s:t}$ is called a time segment of $\bm{x}^i$, where $\llbracket s, t\rrbracket, 1\leq s<t\leq T$ is called a window.
\end{definition}

\begin{definition}\textbf{Target Region}
At the last date $T$, we predict the future incidences for location $i_0\in[1, N]$ beyond $T$. We call $i_0$ the target region and $\bm{x}^{i_0}_{1:T}$ the target time series. 
\end{definition}

\begin{definition}\textbf{Reference Regions}
The regions other than the target region $i_0$ are called reference regions. The reference time series are $\bm{x}^i_{1:T}$ where $i\neq i_0$.  In a generalized definition, reference regions could include the target region. 
\end{definition}

\begin{definition}\textbf{Additional Features}
Besides historical incidences in each region, other features might be available including demographic information, mobility index, and interventions. For each region $i$, time-independent features are concatenated into a single vector $\bm{u}^i$, and time-dependent ones into another time series $\bm{r}^i_t$.
\end{definition}

\noindent \textbf{Problem Statement} Given $N$ time series $X^i_{1:T}$ ($i\in[1, N]$) and additional features, we aim to predict future incidences in a target region $i_0\in[1, N]$ over $H$ consecutive days after $T$, i.e. $\bm{x}^{i_0}_{T+1:T+H}$. \nop{Specifically, when $G=0$, we forecast the next $H$ days after the last observed date $T$. Sometimes, it is more practical to directly forecast incidences $G$ days ahead.}

\begin{figure*}[h]
    \centering
    \includegraphics[scale=0.4]{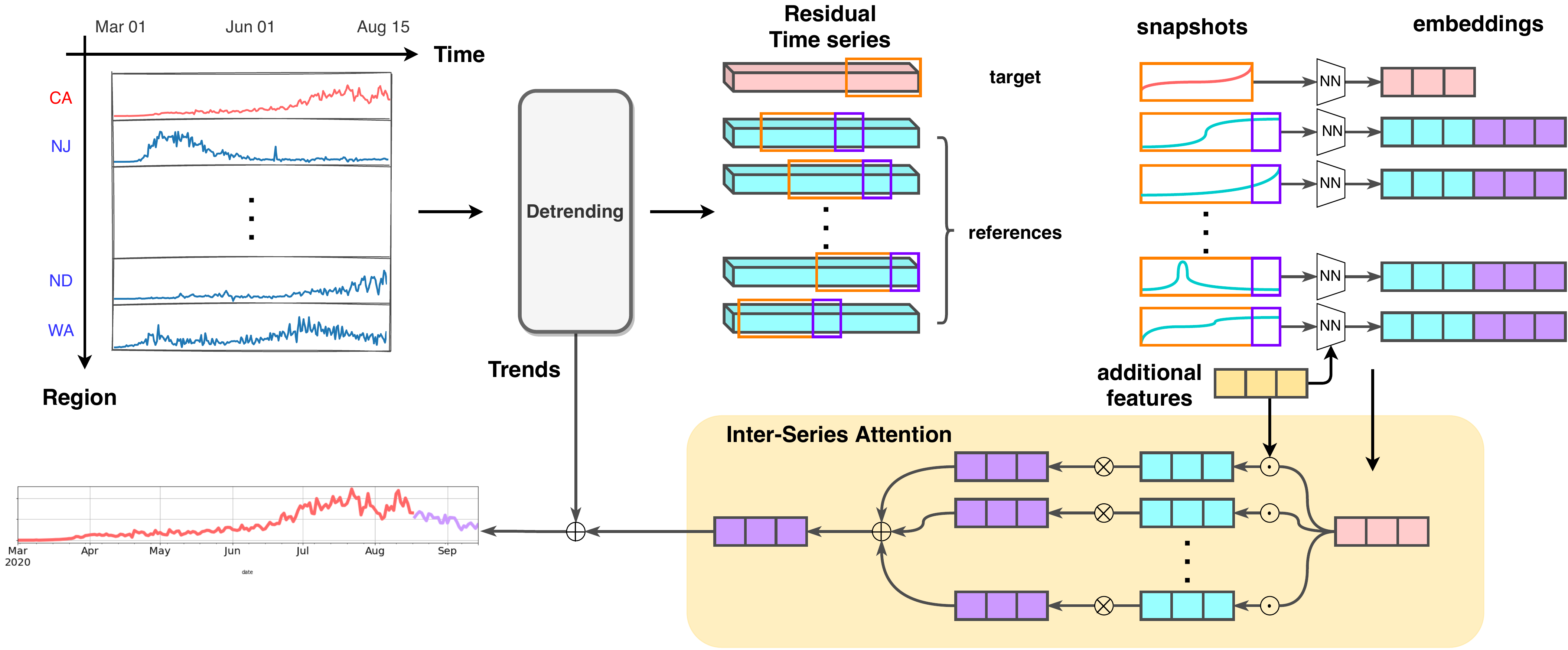}
    \caption{Our proposed Inter-series Attention Network. Best view in color.}
    \label{fig:arch}
\end{figure*}

\section{Methodology}\label{sec:method}
Traditionally, epidemic forecasts are made by analyzing only the growth pattern of incidences.  \cite{balcan2009multiscale, balcan2010modeling, deng2019graph, gao2020stan} take the incidences from neighboring regions into consideration as diseases spread through social interaction. Rather than explicitly modeling the disease spreading process, we take a bold step to directly compare the incidence curves across regions. Once the similarities between the current incidences in the target region with the past time segments in reference regions are identified, the following incidences in the reference regions can be used to forecast the future incidences in the target region.\nop{On the other hand, when the intervention policy changes, the feature vector $\bm{u}^i$ shall have a way to influence the forecasting.} Hence, the critical challenge in implementing our idea is to (1) define a representation of a time segment; (2) identify similar segments in reference regions through the representations; and (3) aggregate their following incidences for forecasting. 
 
%Instead of using predefined spatial features to explicitly model disease spreading process, we propose to find correlation between different regions by comparing growth patterns in a period of time directly. Intuitively, since COVID-19 outbreak in some coastal states, such as New York, New Jersey and Washington, took place before other inland states, and similar development of public health responses were seen under various level of intervention, the forecasting of a target region can refer to the historical incidence in some reference regions. At a high level, we leverage similarities between the currently observed stage in the target region with the past stages in the reference regions. Provided with the assumption that the current intervention will continue, we can expect that the future stages in the target region be close to the following stages of the similar reference stages in the past. Hence, the critical challenge in implementing our idea is to find 1). the reference regions, 2). the past periods similar to the current target stage in a reference region.

%\from{xifeng}{can f, and g be found in the remaining descriptions?  I only found g. not sure if they are the same. }

Formally, we introduce an embedding function $\phi(\cdot)$ to encode a time series segment $\bm{x}_{t-l+1:t}$ into a vector, and then use dot-product of vectors to measure similarity. The following incidences $\bm{x}_{t+1:t+h}$ is also encoded by another embedding function $\psi(\cdot)$ for further aggregation. However, while there are comparable short-term patterns that can be extracted from time series segments, there are also non-stationary long-term trends that hinder reasonable comparison and aggregation of local patterns within segments. \nop{Hence heuristic trend filtering methods based on smoothing regularization \cite{kim2009ell_1} are not applicable.}

We resolve the problem in two steps. First, we apply a trainable detrending module to the raw time series to remove long-term trends so that incidences across different regions are more comparable.  Second, we take rolling windows from residual time series and transform them into a common feature space using normalized convolution as embedding functions $\phi(\cdot)$ and $\psi(\cdot)$. The embedding of the recent window in the target region is then compared with windows from references to produce weights for combining the following incidences of each reference window.  In such pairwise comparisons, differences in both time-dependent and time-independent features are taken into account so that the curves in corresponding windows can be better aligned. The combinations are then added to the extrapolation of filtered trends to generate the final prediction. We jointly train both modules in an end-to-end manner so that both the long- and short-term patterns can be decoupled in an adaptive way.

Figure \ref{fig:arch} gives an overview of the framework. In the following subsections, we introduce each component in details.

%\from{xifeng} {It seems you do not use l0,  is it a0?} 

\subsection{Detrending}
We adopt a learnable Holt smoothing model (\cite{smyl2020hybrid}) to remove long-term trends from the raw time series. Specifically, we introduce a set of parameters $\theta^i_e = [a^i_0; b^i_0;\alpha^i;\beta^i]$ per series, where $a^i_0$ is the initial level, $b^i_0$ is the initial trend, $\alpha^i$ is the level smoothing coefficient and $\beta^i$ is the trend smoothing coefficient. Then Holt's equations (\cite{holt2004forecasting}) are launched to iteratively derive levels and piecewise linear slopes in $\bm{x}^i_{1:T}$,
\begin{equation}
\begin{aligned}
    a^i_t &= \alpha^i\bm{x}^i_t + (1-\alpha^i)(a^i_{t-1}+b^i_{t-1}),\\
    b^i_t &= \beta^i(a^i_t - a^i_{t-1}) + (1-\beta^i)b^i_{t-1},\\
    \bm{\hat{x}}^i_t &= \bm{x}^i_t - a^i_t .
\end{aligned}
\end{equation}
After detrending, the residual time series $\bm{\hat{x}}_{1:T}$ will contain  short-term patterns for further processing.  Projection from the long-term trend is generated by simple linear extrapolation, 
\begin{equation}
\begin{aligned}
    \bm{\bar{x}}^i_{t+h} = a^i_t + h b^i_t .
\end{aligned}
\end{equation}
A more sophisticated detrending process might further boost performance; we leave it for future study.  The detrending process is applied to all the time series and the residual time series are fed into the following attention module.

\subsection{Attention Module}
As COVID-19 is a new disease, we do not have its historical data in the past seasons. Hence, it is critical to leverage limited data from the same season, but across different regions, i.e. model the correlations between regions that have been undergoing the pandemic.  Without detailed information about spatial dynamics such as population movement, we instead employ attention mechanism to measure the relation of one region to other regions by directly comparing the incidence curves after trend filtering. Since there are many stages in a dynamical epidemiological process, it is necessary to learn a representation for each time period in a region for alignment in attention. In light of this idea, we apply a convolution layer to encode the residual time series segment $\bm{\hat{x}}_{t-l+1:t}$ to a vector, based on which attention scores measuring similarity between regions are computed. \nop{ Finally the forecast for the target region is produced from a combination of succeeding reports of similar stages in reference regions.}
 
\subsubsection{Segment Embedding}
Even after detrending, the scales of reported numbers in residual time series are still quite different across regions.  It is important to normalize residuals before embedding. We empirically find it better to apply min-max normalization to the cumulative sum of incidence time series, which can be regarded as a kind of smoothing. Specifically, for a rolling window of size $l$ representing a period of time, i.e. $\bm{\hat{x}}^i_{t-l+1:t}, t\in[l, T]$, we compute its cumulative sums and apply the min-max normalization to the monotonically increasing series, 
\begin{equation}\label{eq:minmax}
    \bm{c}^i_j = \sum_{k=t-l+1}^j \bm{\hat{x}}^i_k; \qquad
    \bm{\tilde{c}}^i_j = \frac{\bm{c}^i_j - \bm{c}^i_{t-l+1} }{\bm{c}^i_t - \bm{c}^i_{t-l+1}},
\end{equation}
for $j\in[ t-l+1,t]$. As a result, the first and last values of the normalized series will consistently be $0$ and $1$ respectively.

We then instantiate the function $\phi(\cdot)$ using a convolution layer with $d$ feature maps to the scaled segment and time-dependent features. The kernel size is empirically selected, and when it is smaller than $l$, average pooling is applied in order to reduce a sequence to a vectorized embedding,
\begin{equation}
    \bm{p}^i_t = \text{AvgPool}\left(\text{Conv}\left(\left[\bm{\tilde{c}}^i_{t-l+1:t};\bm{r}^i_{t-l+1:t}\right]\right)\right)\in\mathbb{R}^d.
\end{equation}
These segment embeddings are used to model similarity in different temporal periods across different regions. 

%\from{xifeng}{we need to give a name.  I call it development embedding so that we can precisely refer it.  }
Likewise, we employ another convolution-pooling layer as $\psi(\cdot)$ to encode the following incidences over $H$ days after each segment into so-called \textbf{development embedding},
\begin{equation}\label{eq:val}
    \bm{g}^i_t = \text{AvgPool}\left(\text{Conv}\left(\bm{\tilde{c}}^i_{t+1:t+H}\right)\right)\in\mathbb{R}^d.
\end{equation}
They represent the succeeding development after encoded segments and will be the references for the prediction of the given target region. In fact, we can pair the segments and references by aligning the time indices, i.e. $\{\bm{p}^i_t, \bm{g}^i_t\}$ for $t\in [l,T-H]$.

\subsubsection{Inter-series Attention}\label{subsec:attention}
Given the embeddings, we use dot-product attention to compare segments and combine the values. Specifically, we linearly map the segment embeddings to query vectors $\bm{q}^i_t$ and key vectors $\bm{k}_i^t$, from which the similarity score is computed.  The development embeddings are projected to value vectors $\bm{v}^i_t$. On the other hand, the additional time-independent features $\bm{u}^i$ are also incorprated into queries and keys.
\begin{equation}
\begin{aligned}
    &\bm{q}^i_t = \bm{W}_Q\bm{p}^i_t + \bm{W}_{u,q}\bm{u}^i;\\
    &\bm{k}^i_t = \bm{W}_K\bm{p}^i_t + \bm{W}_{u,k}\bm{u}^i;\\
    &\bm{v}^i_t = \bm{W}_V\bm{g}^i_t;\\
\end{aligned}
\end{equation}
For a target region $i_0$, we take $\bm{q}^{i_0}_T$ for the last segment and compute its similarity with all the keys from other time segments across all the regions, which is then used to obtain a weighted sum of values. 
\begin{equation}\label{eq:attn}
    \bm{\hat{v}}^{i_0}_T = \sum_{i,t\in\Omega}\frac
    {\exp\left(\langle\bm{q}^{i_0}_T, \bm{k}^i_t\rangle\right)}
    {\sum_{i,t\in\Omega}\exp\left(\langle\bm{q}^{i_0}_T, \bm{k}^i_t\right)}\bm{v}^i_t,
\end{equation}
where $\Omega = [1,N] \times [l,T-H]$. In this way, the past observations in both the target region and reference regions are fully utilized. 
% Different from self-attention in common sequence modeling tasks \cite{vaswani2017attention}, the attention in our model is distributed across multiple sequences, which takes $O(N^2T^2)$ steps of comparison and aggregation. In that case, we propose to reduce the complexity by only taking the most similar rolling window from each region into account.
% \begin{equation}
% \begin{aligned}
%     \bm{\omega}^i &= \max_t \left\{\langle\bm{q}^{i_0}_T, \bm{k}^i_t\rangle + \langle\bm{z}^{i_0}_T, \bm{z}^i_t\rangle\right\}\\
%     \bm{\hat{v}}^{i_0}_T &= \frac{\sum_{i=1}^N\bm{v}^i_t\exp(\bm{\omega}^i)}{\sum_{i=1}^N\exp(\bm{\omega}^i)}
% \end{aligned}
% \end{equation}
% Empirically, we find this simple strategy introduce minimum performance degradation or even some improvement.
The weighted combination of values $\bm{\hat{v}}^{i_0}_T$ is then linearly projected to an estimate of $\bm{\tilde{c}}^{i_0}_{T+1:T+H}$. We apply the inverse transformation of \eqref{eq:minmax} to get an estimate of $\bm{\hat{x}}^{i_0}_{T+1:T+H}$, denoted by $\bm{\hat{y}}^{i_0}_{T+1:H}$.

In the end, the estimate from attention module is added to the extrapolations in the detrending module to produce the final forecast $\bm{y}^{i_0}_{T+1:T+H}$, where
\begin{equation*}
    \bm{y}^{i_0}_t = \bm{\bar{x}}^{i_0}_{t} + \bm{\hat{y}}^{i_0}_{t}, \qquad t\in[T+1,T+H]. 
\end{equation*}
% \paragraph{Regression}
% When forecasting a type of incidence, it is helpful to incorporate other incidences simultaneously. For example, the incidence hospitalizations come from recent new confirmed cases, and incidence deaths take place in hospitalized patients. In light of this, we also introduce a regression model

\subsection{Joint Training}
The model can be trained by minimizing the joint loss with respect to the parameters in all the modules. The joint loss is an aggregation of prediction error $E(\cdot,\cdot)$ computed in two steps. First, for a single target region, we compare our forecasts and ground truths for different $T$, i.e. lengths of history. Second, we take the aggregated loss in the first step for every region. Formally, the joint loss is defined as
\begin{equation}
    \mathcal{L} = \sum_{i=1}^N\sum_{T=l}^{L-H} E(\bm{y}^i_{T+1:T+H}, \bm{x}^i_{T+1:T+H})
\end{equation}
where $L$ is the total number of available historical reports, and $l$ is the minimum required history length. In our experiments, we choose Mean Absolute Error (MAE) to be the error metric $E(\cdot,\cdot)$, i.e.
\begin{equation*}
    E(\bm{y}^i_{T+1:T+H}, \bm{x}^i_{T+1:T+H}) = \frac{1}{H}\sum_{t=T+1}^{T+H}\vert\bm{y}^i_{t} - \bm{x}^i_t\vert
\end{equation*}

\section{Experiments}
In this section, we demonstrate the effectiveness of the proposed model on real COVID-19 datasets. We intend to answer the following questions:
\begin{itemize}
    \item Can \textbf{ACTS} outperform the popular COVID-19 forecasters referred at CDC and other state-of-the-art deep learning models?
    \item How much does each component of \textbf{ACTS} contribute to the model performance?
    \item What kind of similarity can inter-series attention capture?
\end{itemize}

\subsection{Experimental Settings}
\paragraph{Dataset}
The COVID-19 incidence data is publicly available at  JHU-CSSE\footnote{\url{github.com/CSSEGISandData/COVID-19}} and COVID tracking project\footnote{\url{covidtracking.com/}}. Additional features are also publicly available \footnote{\url{github.com/descarteslabs/DL-COVID-19}} \footnote{\url{github.com/djsutherland/pummeler}} \footnote{\url{data.world/liz-friedman/hospital-capacity-data-from-hghi}}. The features we used include total population, population density, ratios of age/gender/race, available hospital beds, and traffic mobility, which are proven to bring marginal accuracy gain in the hospitalization forecasting task in our experiments. \nop{The main accuracy gain \ACTS\ is not from the use of these additional features.}  The dataset covers the reports up to September 27, 2020 from 50 states and DC in the US. 

\paragraph{Evaluation Protocol}
As required by CDC, we predict the incidence data over the next 4 weeks at a given date and compare the forecasts with the reported ground truths. Suppose we are predicting the new confirmed cases in the state of California starting from 08/16. As context, we are provided a daily time series consisting of incidences in all the states till 08/15.\nop{Meanwhile, we have increased cases from other US states up to 08/15 as well. Moreover, time-dependent data such as mobility measures up to 08/15 and time-independent data such as demographic statistics in all states are available too. Based on these contexts, we aim to predict either the incident cases from 08/16 to 09/12 on a daily basis for hospitalizations, or the aggregated increased cases of the $4$ weeks covering the same period, i.e. 08/16-08/22, 08/23-08/29, 08/30-09/05, 09/06-09/12, for new cases and deaths.} 
There are three forecasting tasks: daily forecasts for new hospitalizations, weekly forecasts for new confirmed cases and deaths. \nop{The forecasting time period is 4 weeks}

The forecasting performance is evaluated in terms of Weighted Absolute Percentage Error (WAPE), defined by the ratio of Mean Absolute Error (MAE) and mean value of ground truths and frequently used in research \cite{li2019enhancing, lim2019temporal}.
%\from{xifeng}{Is is popularly used. citation here.} 
At each prediction date, we keep the data in the last $7$ days for validation, and the remaining historical data for training. We use the validation data to tune the hyperparameters and to avoid overfitting by early stopping. Other implementation details can be found in Appendix too.

\paragraph{Baselines}
We compare the performance of the epidemic models featured at CDC, including
\begin{itemize}
    \item \textbf{YYG}~\cite{Gu2020Covid}: An SEIR model with learnable parameters that attracts a lot of attention from media;
    \item \textbf{CU}~\cite{pei2020initial}: A metapopulation SEIR model developed by researchers in Columbia University;
    \item \textbf{UCLA}~\cite{zou2020epidemic}: An SuEIR model using machine learning developed by Statistical Machine Learning Lab at UCLA;
    \item \textbf{ERDC}\footnote{\url{https://github.com/reichlab/covid19-forecast-hub/blob/master/data-processed/USACE-ERDC_SEIR/metadata-USACE-ERDC_SEIR.txt}}: An SEIR model that considers unreported infections and isolated population developed by US Army Engineer Research and Development Center;
    % \item \textbf{IHME}~\cite{covid2020forecasting}: Combination of a mechanistic disease transmission model and a curve-fitting approach developed by Institute of Health Metrics and Evaluation;
    \item \textbf{LANL}~\cite{lanl2020covid}: A statistical dynamical growth model accounting for population susceptibility developed by Los Alamos National Laboratory;
    \item \textbf{CovidSim}\footnote{\url{https://www.covid19sim.org/documents/outbreak-methods}}: Machine learning model based on generalized random forests.
\end{itemize}
The first four are compartmental models and the last two rely on statistical modelling. Other than these conventional models, we also evaluate three deep learning models for time series forecasting,
\begin{itemize}
    \item \textbf{DeepCOVID}~\cite{rodriguez2020deepcovid} An operational deep learning framework designed for real-time COVID-19 forecasting developed by Georgia Tech;
    \item \textbf{ConvTrans}~\cite{li2019enhancing} A self-attention based Transformer model that also employs convolutions for pattern representations;
    \item \textbf{TFT}~\cite{lim2019temporal} A self-attention based deep learning model with feature selection.
\end{itemize}
%\from{xifeng}{Need to explain why a single layer attention, otherwise, better uses the original implementation}
We implement the ConvTrans and TFT and tune the hyperparameters using the validation data. All of our implementations run on a server with an Intel i7-6700K CPU and a single GTX 1080Ti GPU. For other baselines, since their implementations are not open-sourced, we take their forecasts submitted to the challenge hosted by CDC \footnote{\url{https://github.com/reichlab/covid19-forecast-hub}}.

\begin{table*}[h]
\begin{tabular}{c|c|c|c|c|c|c|c|c|c|c|c}
\toprule
\multirow{2}{*}{\textbf{}}  & \multirow{2}{*}{\textbf{}} & \multicolumn{9}{c}{\textbf{Method}}   \\
\cmidrule{3-12}
                       &           & YYG      & CU       & UCLA   & ERDC   & LANL   & Covid    & Deep      & Conv      & TFT    & ACTS   \\
                       &           &          &          &        &        &        & Sim      & COVID     & Trans     &        &    \\
                       
\midrule                            
\multirow{3}{*}{06/21} & C         & -        & -        & -      & -      & 0.51   & -        & -         & 1.09      & 0.51   & \bf0.39$\pm$0.01 \\
                       & H         & -        & 1.91     & -      & -      & 1.08   & 0.95     & \bf0.63   & 1.22      & 0.80   & 0.80$\pm$0.02 \\
                       & D         & 0.52     & 1.48     & 0.56   & -      & 0.58   & 1.46     & 0.66      & 1.09      & 0.67   & \bf0.45$\pm$0.01 \\
\midrule                              
\multirow{3}{*}{07/05} & C         & -        & -        & -      & -      & 0.37   & -        & -         & 0.37      & 0.39   & \bf0.33$\pm$0.01 \\
                       & H         & -        & 0.98     & 1.23   & 0.66   & 0.95   & -        & 0.65      & 1.08      & 0.84   & \bf0.61$\pm$0.04 \\
                       & D         & 0.45     & 0.65     & 0.53   & \bf0.38& 0.52   & -        & 0.85      & 0.60      & 0.51   & 0.60$\pm$0.01 \\
\midrule                            
\multirow{3}{*}{07/19} & C         & -        & -        & -      & -      & \bf0.27& -        & -         & 0.50      & 0.44   & 0.31$\pm$0.01 \\
                       & H         & -        & 0.67     & 1.24   & 0.77   & 0.78   & 1.71     & 0.70      & 0.99      & 0.66   & \bf0.60$\pm$0.03 \\
                       & D         & 0.30     & 0.43     & 0.39   & 1.10   & 0.48   & 0.33     & 0.4506    & 0.54      & 0.67   & \bf0.28$\pm$0.01 \\
\midrule                            
\multirow{3}{*}{08/02} & C         & -        & -        & -      & -      & 0.30   & -        & -         & 0.24      & 0.24   & \bf0.16$\pm$0.04 \\
                       & H         & -        & 0.67     & 0.95   & 0.71   & 0.68   & 1.66     & 0.79      & 0.93      & 0.92   & \bf0.66$\pm$0.09 \\
                       & D         & 0.24     & 0.37     & 0.27   & 0.57   & 0.44   & 0.26     & 0.29      & 0.45      & 0.38   & \bf0.21$\pm$0.01 \\
\midrule                            
\multirow{3}{*}{08/16} & C         & -        & 0.67     & 0.35   & 0.28   & 0.29   & 0.23     & -         & 0.33      & 0.55   & \bf0.20$\pm$0.03 \\
                       & H         & -        & 0.64     & 0.99   & 0.60   & 0.65   & 1.38     & 0.98      & 0.96      & 0.92   & \bf0.57$\pm$0.02\\
                       & D         & \bf0.19  & 0.42     & 0.25   & 0.53   & 0.34   & 0.27     & 0.28      & 0.44      & 0.31   & 0.23$\pm$0.01 \\
\midrule                            
\multirow{3}{*}{08/30} & C         & -        & 0.43     & 0.31   & 0.34   & 0.33   & 0.23     & -         & 0.36      & 0.29   & \bf0.23$\pm$0.03 \\
                       & H         & -        & 0.66     & 0.91   & 0.68   & 0.69   & 1.31     & 0.83      & 0.93      & 0.82   & \bf0.58$\pm$0.03 \\
                       & D         & \bf0.20  & 0.41     & 0.23   & 0.56   & 0.34   & 0.25     & 0.36      & 0.42      & 0.40   & 0.25$\pm$0.02 \\
\bottomrule
\end{tabular}
\caption{Forecasting performances across different time periods for different types of incidence data in terms of WAPE. A smaller value indicates better performance. We also include the variance of our model's performance by running 5 times with different random initalizations. ``-'' means the forecasting results of the corresponding baseline are not available.}
\label{tab:results}
\end{table*}

\subsection{Performance Comparison}
Table \ref{tab:results} shows the forecasting performance on 6 different dates. Three types of incidence data, namely confirmed cases (C), hospitalizations (H) and deaths (D) are separately predicted. We have three key observations: (1) In 13 out of 18 cases, \textbf{ACTS} outperforms other algorithms by a considerable margin. On average, it improves 9\%, 5\%  and 4\% over the best of these algorithms for C, H and D, respectively. (2) \textbf{ACTS} is more favorable on recent days when there are more abundant data available, showing that data-driven methods benefit from more data. (3) The two deep learning approaches ConvTrans and TFT do not exhibit strong performance.  The main difference between ours and these approaches is the employment of attention across multiple time series, which dramatically boosts the performance. Note that our model can be trained in less than 5 minutes and inference takes only seconds.

\subsection{Ablation Study}
For deeper understanding of our model, we disable each component of \textbf{ACTS} to examine its contribution:
\begin{itemize}
    \item \textbf{ACTS-d} We remove the detrending module and obtain an attention-only forecaster;
    \item \textbf{ACTS-n} We remove the normalization in segment embedding;
    \item \textbf{ACTS-i} We restrict the attention to the target time series only. The model degenerates to an auto-regressive model similar to ConvTrans and TFT;
    \item \textbf{ACTS-f} We remove the additional features in the model and only rely on incidence data.
\end{itemize}

\begin{figure}[t]
    \centering
    \includegraphics[scale=0.38]{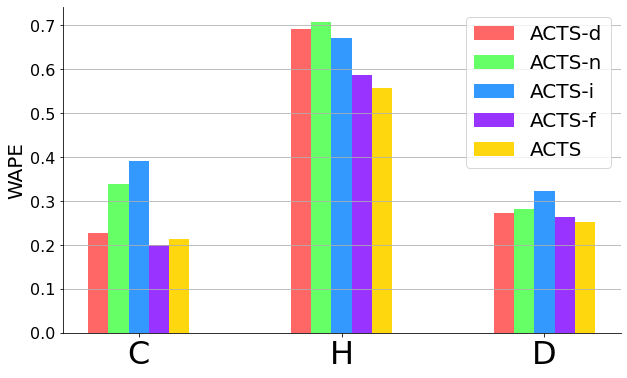}
    \caption{Empirical effects of each component of \textbf{ACTS} on forecasting error.}
    \label{fig:abl}
\end{figure}

The hyperparameters of all variants are kept the same. We compare their performance against \textbf{ACTS} using training data up to August 30, 2020.  Figure \ref{fig:abl} depicts the results, based on which we have the following observations:
\begin{itemize}
    \item Overall every component of \textbf{ACTS} has positive effects on forecasting accuracy, except that the introduction of additional features has mixed effect.  We suspect that either better modelling could help or their effect has been absorbed by the incidence time series;
    \item Among all the components, inter-series attention has the most significant impact on the performance, which proves that our design of attention crossing multiple time series is valid. It can capture cross-region similarity in COVID-19 forecasting;
    \item The detrending module makes some contribution.  We believe it has the potential for further improvement, e.g. employing advanced trend filtering or even epidemic models.
\end{itemize}

\subsection{Cross-region Similarity}
A key feature of \ACTS\ is that it can capture similarity between regions via attention from data. According to \eqref{eq:attn}, the reference set $\Omega$ is common for any target regions $i_0$, and the learned attention distribution is determined by $\bm{q}^{i_0}_T$. Hence, we directly take those $d$-dim queries for every region and apply K-means clustering to group them.  In this experiment, we use the death forecasting model as an example, where $T$ is August 30, 2020, and $K=4$ is selected based on the Elbow method \cite{marutho2018determination}.

\begin{figure}
    \centering
    \includegraphics[scale=0.5]{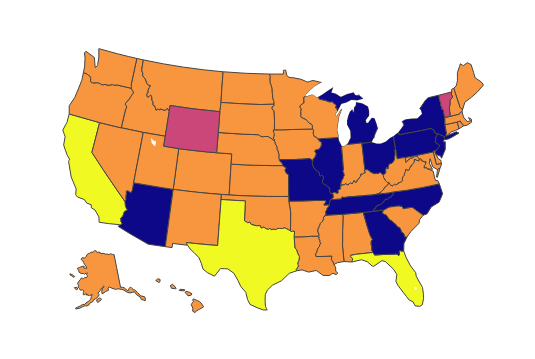}
    \caption{Groups of the US states learned by inter-series attention on death tolls by August 30, 2020.}
    \label{fig:region}
\end{figure}

A colored map is shown in Figure \ref{fig:region} based on obtained clusters.  We can see that California, Texas and Florida, the three states recently hit most seriously are grouped together.  Furthermore, states like Arizona, Illinois, North Carolina and Georgia are recognized since they also suffer severe crisis.  Interestingly, the states of Wyoming and Vermont are distinguished by our model, in which few deaths are observed for a long period. Overall, our method is able to identify similarities between regions to a certain degree.

\section{Conclusion}
In this paper, we present \textbf{ACTS} for COVID-19 forecasting, a purely data-driven framework for an urgent forecasting problem concerning the entire world.  It extends the popular deep learning technique, namely attention mechanism, to learning inter-series similarity for time series forecasting. Above that, we also introduce a detrending component to model long-term trends that are difficult for attention model to capture. Both modules are learned jointly based solely on COVID-19 incidence data and a handful of simple features. Without any domain knowledge, our model can empirically outperform many strong forecasters that are featured by CDC. On the other hand, we find great potential for improvement for trend filtering and incorporating additional features, which is left to future work.
\bibliography{ref}
\bibliographystyle{acm}

\newpage
\appendix
\section{Implementation Details}
We implement our model and its variants using PyTorch. The hyperparameters we used in all of our experiments are listed in the following table.
\begin{table}[h]
    \centering
    \begin{tabular}{c|c}
    \toprule
        Hyperparam & values\\
        \midrule
        hidden size $d$ & [16, 32] \\
        segment length $l$ & [7, 14] \\
        horizon $H$ & 7 \\
        learning rate & [$0.001$, $0.005$, $0.01$]\\
        \# training iterations & [600, 1200, 1800]\\
    \bottomrule
    \end{tabular}
    \caption{Hyperparameters}
    \label{tab:my_label}
\end{table}

Exact values are selected by validation loss. We train all of our implementations using an Intel i7-6700K CPU and a single NVIDIA GTX 1080 Ti GPU (CUDA 10.2) hosted by Ubuntu 16.04. Each training iteration takes approximately $\frac{1}{6}$ second.

Since for each task we need to forecast 4 weeks, we separately predict each week using the same attention module to avoid long-term errors. To predict the $k$-th week,  we replace \eqref{eq:val} by 
$$\bm{g}^i_t = \text{AvgPool}\left(\text{Conv}\left(\bm{\tilde{c}}^i_{t+(k-1)H+1:t+kH}\right)\right), $$ i.e. take the development $(k-1)H$ days after the corresponding segment. For case and death forecasting where forecasts are aggregated by weeks, we directly aggregate $\bm{g}^i_t$ within a week before applying the final transformation.

In generating final prediction, we avoid negative values by clipping partial predictions from both detrending module and attention module.

Our implementation is open-sourced at \url{https://github.com/Gandor26/covid-open}.

\section{Example Forecasts}
We here show example forecasts of hospitalizations and deaths in three representative states, i.e. Florida, Maryland and Virginia, to give a qualitative demonstration. Our forecasts are accompanied by two best baselines in either task.

We can see that in most cases \textbf{ACTS} fits the ground truths better than baselines. An exception is the hospitalization forecast for Maryland, in which \textbf{ACTS} has systematical underestimation. This is because the downward trend captured by the detrending module significantly drags the final prediction down. It indicates that more advanced trend filtering method can further improve the performance of our model.

\begin{figure}[h]
    \centering
    \includegraphics[scale=0.4]{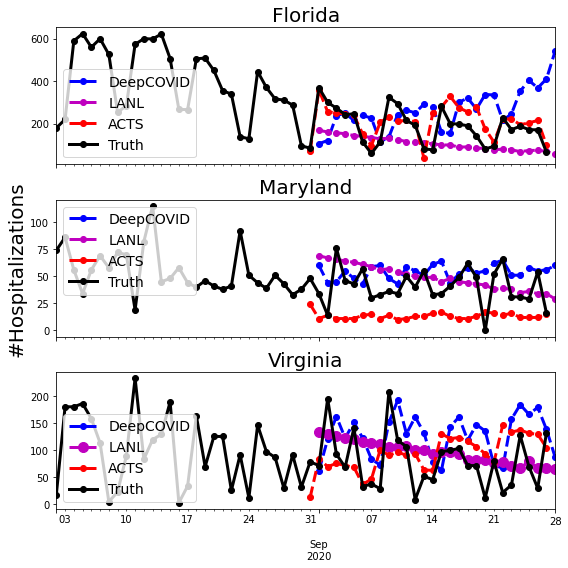}
    \caption{Daily hospitalization forecasts on August 30, 2020}
    \label{fig:hosps}
\end{figure}
\begin{figure}[h]
    \centering
    \includegraphics[scale=0.4]{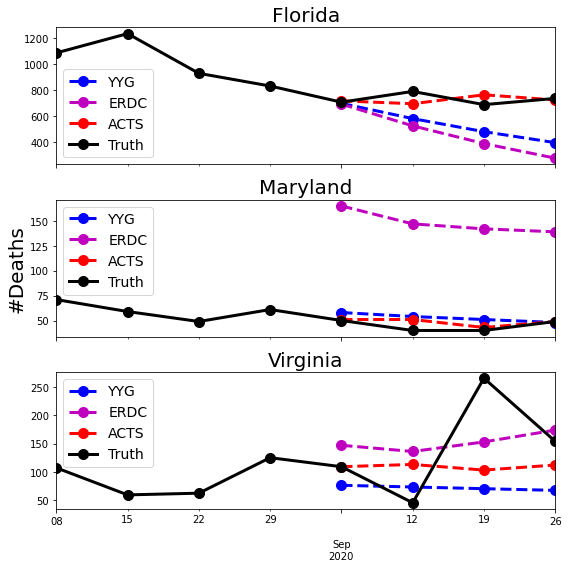}
    \caption{Weekly death forecasts on August 30, 2020}
    \label{fig:death}
\end{figure}

\end{document}